\documentclass[lettersize,journal]{IEEEtran}
\usepackage{amsmath,amsfonts}
\usepackage{algorithmic}
\usepackage{algorithm}
\usepackage{array}
\usepackage{caption}
\usepackage{subcaption}
\usepackage{textcomp}
\usepackage{url}
\usepackage{verbatim}
\usepackage{graphicx}
\usepackage{cite}

\usepackage[utf8]{inputenc} 
\usepackage[T1]{fontenc}    
\usepackage{hyperref}       
\usepackage{url}            
\usepackage{booktabs}       
\usepackage{amsfonts}       
\usepackage{microtype}      
\usepackage{xcolor}         
\usepackage{graphicx}
\usepackage{amsmath,amssymb,amsthm}
\usepackage{algorithm, algorithmic, setspace}

\newcommand{\argmin}[1]{\underset{#1}{\mathrm{argmin\,}}}
\newcommand{\R}{\mathbb{R}}
\newcommand{\reg}{\mathtt{r}}
\newcommand{\Reg}{\mathtt{R}}

\newcommand{\loss}{\mathcal{L}}
\newcommand{\w}{\theta} 
\newcommand{\what}{\hat{\theta}}
\newcommand{\m}{m} 
\newcommand{\diag}{\text{diag}}
\newcommand{\mtrue}{\widetilde{m}} 
\newcommand{\mstar}{m^{\star}} 
\newcommand{\supp}{S} 
\newcommand{\suppc}{\supp_c} 
\newcommand{\yhat}{\hat{y}}

\newtheorem{lemma}{Lemma}

\newtheorem{theorem}{Theorem}

\newtheorem{assumption}{Assumption}

\renewcommand{\qed}{\hfill$\blacksquare$}

\begin{document}

\title{Playing the lottery with concave regularizers\\for sparse trainable neural networks \thanks{Sophie M. Fosson's work is partially included in the project NODES which has received funding from the MUR – M4C2 1.5 of PNRR with grant agreement no. ECS00000036. Andrea Migliorati's work was partially supported by the European Union under the Italian National Recovery and Resilience Plan (NRRP) of NextGenerationEU, partnership on “Telecommunications of the Future” (PE00000001 - program “RESTART”).}}

\author{Giulia Fracastoro, Sophie M. Fosson, Andrea Migliorati, Giuseppe C. Calafiore}
%

\markboth{Journal of \LaTeX\ Class Files,~Vol.~14, No.~8, August~2021}%
{Shell \MakeLowercase{\textit{et al.}}: A Sample Article Using IEEEtran.cls for IEEE Journals}


\maketitle

\begin{abstract}
The design of sparse neural networks, i.e., of networks with a reduced number of parameters, has been attracting increasing research attention in the last few years. The use of sparse models may significantly  reduce the computational and storage footprint in the inference phase. 

In this context, the lottery ticket hypothesis constitutes a breakthrough result, that addresses not only the performance of the inference phase, but also of the training phase. It states that it is possible to extract effective sparse subnetworks, called {\emph{winning tickets}}, that can be trained in isolation. 
The development of effective methods to {\emph{play the lottery}}, i.e., to find winning tickets, is still an open problem.
In this paper, we propose a novel class of methods to play the lottery. The key point is the use of concave regularization to promote the sparsity of a relaxed binary mask, which represents the network topology.

We theoretically analyze the effectiveness of the proposed method in the convex framework. Then, we propose extended numerical tests on various datasets and architectures, that show that the proposed method can improve the performance of state-of-the-art algorithms. 
\end{abstract}

\begin{IEEEkeywords}
 Lottery ticket hypothesis, concave regularization, neural network pruning, sparse optimization
\end{IEEEkeywords}

\section{Introduction}\label{sec:in}
Neural network pruning refers to the sparsification of a neural architecture by removing unnecessary parameters, i.e., either connections (weights) or neurons. As a matter of fact, neural networks are often overparametrized and their size can be significantly decreased while keeping a satisfactory accuracy. Pruning allows us to reduce the computational costs, storage requirements, and energy consumption of a neural network in the inference phase; see, e.g., 
\cite{lecun90,han15,lou18,fra19}.
 This has several advantages, ranging from the circumvention of overfitting \cite{lecun90} to the possibility of implementing deep learning in embedded mobile applications \cite{yang2017designing,mol17}. 

Learning pruned (or sparse) neural networks is a challenging task, which has drawn substantial attention in the last years. In the literature, two main approaches are considered. The classic one is {\emph{dense-to-sparse}}:  the model is dense at the beginning and during the training, while the output of the training is a sparse network. In this context, a common practice is the following three-stage iterative procedure \cite{han15,blalock2020state}: first, the dense method is trained, then the sparse subnetwork is extracted, and finally the subnetwork  is retrained, by starting from the weights of the trained dense model; this last stage is known as {\emph{fine-tuning}}. Most of the proposed dense-to-sparse training methods reduce the number of non-zero parameters (i.e., the $\ell_0$ norm of the parameter vector) by pruning the weights with the largest magnitude \cite{han15,zhu18} or via $\ell_0$ regularization \cite{lou18}. More recently, $\ell_1$ and concave regularizations have been studied and tackled through proximal gradient methods \cite{yan20,yun21}. The benefit of the dense-to-sparse approach is the acceleration of the inference task by using the sparse model; nevertheless, the training remains computationally intense.

The second and more recent approach, known as {\emph{sparse-to-sparse}}, tackles the computational burden of the training phase. Basically, it states that we can learn the sparse architecture and then train it in isolation.
However, learning a sparse topology that can be trained in isolation is very challenging. In particular, it has been observed that retraining a sparse network obtained through a dense-to-sparse method from a random initialization often yields a substantial loss of accuracy. 
A way to circumvent this problem is {\emph{rewinding}} the weights to the original initialization of the dense model, as proposed in \cite{fra19}. More precisely, in \cite{fra19}, the authors conjecture the {\emph{lottery ticket hypothesis}} (LTH): a dense, randomly-initialized neural network $\mathrm{N_d}$ contains a small subnetwork $\mathrm{N_s}$ that achieves a test accuracy comparable to the one of $\mathrm{N_d}$, with a similar number of iterations, provided that $\mathrm{N_s}$ is trained in isolation with the same initialization of $\mathrm{N_d}$.


The effective subnetworks mentioned in the LTH are named  {\it{winning tickets}} as they have won the {\emph{initialization lottery}}. While their existence is proven in \cite{fra19}, their extraction is not straightforward. As a matter of fact, the development of effective algorithms to win the lottery is still an open problem. In \cite{fra19}, a method based on iterative magnitude pruning (IMP) is proposed, which still represents the state-of-the-art heuristic to play the lottery. In \cite{fra20}, IMP is refined to deal with instability: the weights are rewinded to an early iteration $k>0$ instead of the initial $\w_0$ \cite[Section 3]{fra20}. Thus, subnetworks are no longer randomly initialized, and they are denoted as matching tickets \cite[Section 4]{fra20}.
In \cite{sav20}, an $\ell_0$ regularization approach is proposed to search winning tickets, based on continuous sparsification, which outperforms the accuracy of IMP in some numerical experiments. Nevertheless, such enhancement is obtained at the price of repeated rounds, which causes a slower sequential search, as discussed in \cite[Section 5.1]{sav20}. 

In this work, we propose a novel method to play the lottery, i.e., to find sparse neural networks that can be trained in isolation by rewinding. The key steps of our approach are the following: we define a binary relaxed mask, which describes the network topology, and we optimize it by applying a continuous, concave regularization. In particular, we consider  $\ell_1$ (which is also convex) and logarithmic regularizers. The rationale behind continuous regularization is that most of the sparse-to-sparse methods used to play the lottery \cite{fra19,sav20} are $\ell_0$-based, which yields hard decisions. For example, in IMP, the percentage of surviving parameters is priorly set, possibly causing the removal of important weights. Instead, a continuous regularization may yield softer decisions in the pruning process, and, as a consequence, more accurate models. To the best of the authors' knowledge, this is the first sparse-to-sparse method employing an $\ell_1$/concave setting.

As discussed throughout the paper, the use of a relaxed binary mask allows us to train the model via projected gradient descent even though the proposed sparsity-promoting regularization is non-differentiable. On the other hand, we show that strictly concave regularization is particularly efficient when applied on binary variables.

The main contributions of this paper can be summarized as follows. We develop a novel strategy to play the lottery, that leverages the use of sparsity-promoting regularization and relaxed binary masks. Then, we provide theoretical conditions under which our approach guarantees to find an optimal sparse mask, at least in the case of convex loss functions. Finally, we extensively validate the proposed method on various datasets and architectures.
%
The paper is organized as follows. In Section \ref{sec:rel}, we illustrate the background and previous work on neural network pruning and LTH, to contextualize the proposed contribution. In Section \ref{sec:met}, we present the proposed method, which is theoretically analyzed in Section \ref{sec:ana}. Section \ref{sec:exp} is devoted to experiments, comparisons, and discussions. Finally, we draw some conclusions. 

\section{Related work}\label{sec:rel}


In recent years, the literature on neural network pruning and LTH has significantly grown, and a complete overview is beyond our purposes. In this section, we review the main works that, for different motivations, are related to our approach.
In Table \ref{table-met}, we report a list of relevant works and their properties.

\begin{figure}
    \centering
    \includegraphics[width=0.45\textwidth]{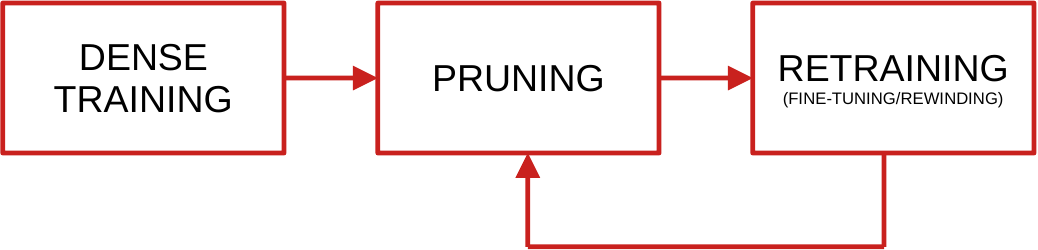}
    \caption{Three-stage pipeline for pruning. The initialization of the retraining stage is done either with fine-tuning or rewinding.}
    \label{fig:3stage}
\end{figure}

\subsection{Dense-to-sparse methods}
Regarding the dense-to-sparse approach, several approaches include a regularization term in the loss function to sparsify the model, the most popular being the $\ell_0$-norm \cite{lou18}, $\ell_1$-norm \cite{yan20} and concave $\ell_p$ norms \cite{yun21}. In these works, training and sparsification are performed in a joint optimization problem. In contrast, in the method proposed in \cite{han15}, training and pruning are separated tasks, and the pruned architecture is iteratively retrained, according to the three-stage pipeline depicted in Fig. \ref{fig:3stage}. The retraining stage is initialized with fine-tuning, i.e., by starting from the parameters obtained in the previous training stage.
Such a three-stage procedure with fine-tuning is very common in the dense-to-sparse approach, see, e.g., \cite{han15,mol17,liu17,liu19}. 

Among more recent works,  in \cite{tar22} a different regularization is proposed, based on the neural sensitivity, to learn sparse topologies with a structure. In \cite{sal22}, an energy-based pruning method is developed, combined with a dropout approach. 
In \cite{ning2020dsa,lin2020hrank,lin2021network,lin2021filter,zha23}, specific methods for filter/channel pruning in convolutional neural networks are developed.

\subsection{Sparse-to-sparse methods}
The above-mentioned three-stage pipeline is also popular in the sparse-to-sparse approach \cite{fra19,sav20}, but with rewinding: in the retraining stage, the parameters are reinitialized by rewinding them to the original initialization; see Section \ref{sec:in}.  
In the IMP method \cite{fra19}, which is by far the most common sparse-to-sparse approach, pruning is performed by removing the $p\%$ parameters with the smallest magnitude. This approach has some drawbacks. As $p$ is priorly set, this may result either in an excessive sparsification with the removal of important weights or in an insufficient sparsification that retains superfluous parameters. Moreover,  very similar weights may be either retained or removed to fulfill $p\%$, which contradicts the principle of saving the most relevant parameters. In \cite{sav20}, the hard effect of magnitude pruning is mitigated by a continuous relaxation of the $\ell_0$ regularization. However, as discussed in Section \ref{sec:in}, in practice this does not outperform IMP. 

In this paper, we tackle these issues by introducing $\ell_1$/$\log$ regularizers to induce sparsity. This results in a softer approach where the output of the optimization/training stage is not expected to be an exact binary mask, but a relaxed mask that can be used for pruning the dense network by setting a suitable threshold $\alpha\in [0,1]$. Also, the dense-to-sparse approach presented in \cite{han15} proposes to set a threshold instead of removing a fixed percentage of the weights. However, in \cite{han15} the threshold is set on the weights of the parameters, which basically can assume any real value and whose range may vary at each layer of the network. Therefore, setting such a threshold may be critical and it requires some prior knowledge of the range of the parameters. In addition, defining a unique threshold for the entire network can be troublesome because weights from different layers might have different orders of magnitude.

\begin{table}[t]
\caption{Classification of main works on neural network pruning. Notation: D2S = dense-to-sparse approach; S2S = sparse-to-sparse approach; 3S = methods exploiting a three-stage iterative procedure, as in Fig. \ref{fig:3stage}. 3S methods may reinitialized either with fine-tuning (FT) or rewinding (RW).\label{table-met}}
{\small{
 \begin{tabular}{l|c|c|c}
 Article - Approach & D2S & S2S & 3S\\
 \hline
 \cite{han15} - Magnitude pruning & \checkmark & & \checkmark {\tiny{FT}}\\
  \cite{lou18} - $\ell_0$ regularization & \checkmark & & \\
 \cite{yan20} - $\ell_1$ regularization    & \checkmark & & \\
 
 \cite{yun21} - $\ell_p$ regularization, $p\in[0,1]$   & \checkmark & & \\
 \cite{tar22} - sensitivity regularization   & \checkmark & & \\
 \cite{sal22} - energy-based dropout   & \checkmark & & \\
 \hline
  \cite{fra19} - IMP with rewinding& & \checkmark & \checkmark {\tiny{RW}}\\
  \cite{sav20} - $\ell_0$ regularization & & \checkmark & \checkmark {\tiny{RW}}\\
  This work - $\ell_1$/log regularization & & \checkmark & \checkmark {\tiny{RW}}\\
\end{tabular}}}
\end{table}

\section{Proposed method}\label{sec:met}
In this section, we present the proposed method.

Let us consider a neural network $f(x; \w)$, where $x$ represents the input data and $\w\in\R^d$ are the weights. Let $\odot$ be the component-wise product between vectors. According to \cite{fra19}, playing the lottery consists of searching a binary mask $\m\in\{0,1\}^{d}$ with $\left\|\m\right\|_{0}\ll d$ such that $f(x; \m\odot \w)$ is a winning ticket, i.e., it achieves performance comparable to $f(x; \w)$ when trained in isolation. 

 The seminal IMP search algorithm proposed in \cite{fra19}  consists of an iterative three-stage pipeline, as illustrated in Fig. \ref{fig:3stage}. The retraining phase is performed by rewinding the parameters to $\m_t\odot\w_0$, where $\w_0$ is the original initialization and $\m_t$ is the current estimated mask.

The proposed method is as follows. We consider the mask $m$ as an optimization variable and we use concave regularization to promote its sparsity. To avoid mixed-integer problems, we relax the binary mask and we consider it as a continuous variable
\begin{equation}
\m\in [0,1]^d.
\end{equation}
Then, we jointly train and sparsify the network by solving 
\begin{equation}\label{eq:theproblem}
\min_{\w\in\R^d, \m\in[0,1]^d} \loss(x; \m\odot \w)+\lambda\Reg(\m)
\end{equation}
where $\Reg$ is a sparsity-inducing, concave regularizer. This approach returns both a sparse relaxed mask and the trained weights. The relaxed mask in $[0,1]^d$ can be interpreted as a relevance score or soft decision for each parameter. To refine these soft results, we prune the parameters with mask value below a small threshold $\alpha\in (0,1)$, $\alpha \ll 1$. We remark that this is not a harsh pruning, which would contradict the proposed soft approach; instead, it is just a way to remove small weights expected to converge to zero.
Furthermore, we notice that setting a threshold in $(0,1)$ is more straightforward than setting a magnitude threshold on the parameters' values as in \cite{han15}, because it does not require information about the parameters' range.

Then, we iterate the procedure as in Fig. \ref{fig:3stage}, with a rewinding strategy, i.e., the three-stage external loop is analogous to the one of IMP.

\begin{algorithm}
\setstretch{1.5}
 \renewcommand{\algorithmicrequire}{\textbf{Input:}}
  \caption{Soft mask pruning with concave regularization}\label{alg}
  \begin{algorithmic}[1]
  \REQUIRE $\w_0\in\R^d$, $m_0=\frac{1}{2}(1,1,\dots,1)\in\R^d$,  $\alpha\in (0,1)$, $T$
  \FORALL{$t=1,\dots,T$}
  \STATE Initialization:\\ $\w=\w_0$, $\m=\m_{t-1}$
  \STATE Optimization:\\ $\min\limits_{\w\in\R^d, \m\in[0,1]^d} \loss(x; \m\odot \w)+\lambda\Reg(\m) ~~\to~~ (\w_t,\m_t)$
  \STATE Pruning:\\ for each $i=1,\dots,d$, if $m_{t,i}<\alpha$, then $m_{t,i}=0$
  \ENDFOR
  \end{algorithmic}
\end{algorithm}

The thorough procedure is summarized in Algorithm \ref{alg}.

To complete the illustration of the algorithm, we introduce the family of considered regularizers $\Reg(\m).$

\subsection{Sparsity-inducing concave regularizers}
%
%
The rationale behind concave regularization is as follows. While sparsity is represented by the $\ell_0$-norm, its use as a regularizer is critical due to non-continuity and non-convexity. For this motivation, the $\ell_1$-norm is often used, as it is the best convex approximation of the $\ell_0$-norm, see \cite{tib96}. In the presence of a convex cost function, by using the $\ell_1$-norm, we recast the overall problem into convex optimization. A typical example is Lasso \cite{tib96}.

On the other hand, continuous concave regularizers have been observed to be more effective than $\ell_1$ regularization, even though they introduce non-convexity in the problem, as their shape is closer to the $\ell_0$-norm.

In this work, we consider concave regularizers $\Reg(\m)$ with the following property.
\begin{assumption}\label{ass:reg}
$\Reg(\m)$ is any function 
\begin{equation}
 \Reg(\m)=\sum_{i=1}^d\reg(\m_i),~~~\m_i\in[0,1]
\end{equation}
such that $\reg:[0,1]\to [0,1]$ is continuous and differentiable in $(0,1)$, concave, non-decreasing, and its image is $[0,1]$.
\end{assumption}
In the literature of signal processing and sparse optimization, the most popular regularizers satisfying Assumption \ref{ass:reg} are
\begin{itemize}
    \item $\ell_1$: $~\reg_{1}(m_i)=\m_i$, see  \cite{tib96,yan20,yun21};
    \item  $\log$: $~\reg_{\epsilon}(m_i)=\frac{\log\left(\frac{\m_i+\epsilon}{\epsilon}\right)}{\log\left(\frac{1+\epsilon}{\epsilon}\right)}$ for any $\epsilon>0$, see \cite{can08rew};
\end{itemize}
Other possible choices are $\ell_q$ norm, see \cite{woo16,yun21}, and minimax concave penalties, see \cite{zha10MCP}.
In this work, we focus our attention on $\ell_1$ and $\log$ regularizers.

In the literature, the use of strictly concave regularization has arisen in the context of linear regression and compressed sensing, see, e.g., \cite{can08rew,fou09,woo16,sel17,fox20}. Then, it has been extended to several machine learning models; we refer the interested reader to the survey \cite{wen18}.
When the cost function is strictly convex, adding a strictly concave regularizer may keep the problem in the convex optimization framework, see, e.g., \cite{bay16}. 
In contrast, the problem is more challenging when the cost function is non-convex. This case is theoretically analyzed, e.g., in \cite{gon13,bre15,hon16}, where much attention is devoted to proving the convergence of the applied algorithms (proximal methods and alternating direction method of multipliers).

Within the family of non-convex cost functions, the case of neural networks is even more difficult. In fact, in deep learning, gradient-based algorithms are commonly used for training, which is in conflict with non-differentiable regularization as in Assumption \ref{ass:reg}, as discussed in Sec. \ref{sub:train}. 


\subsection{Discussion on the training algorithm}\label{sub:train}
The training phase of the proposed approach requires locally solving \eqref{eq:theproblem}. In principle, the $\ell_1$/ $\log$ regularization is critical for the application of gradient-based training algorithms, due the non-differentiability in zero. However, in our approach we regularize $\m\in [0,1]^d$, thus we can use a projected gradient-based algorithm, without differentiability issues.

We remark that $\ell_1$ regularization is popular in deep learning and it is usually implemented in libraries such as TensorFlow and PyTorch. However, since $\ell_1$ norm is not differentiable at zero, it is usually implemented with a subgradient approach, namely the gradient of $|x|$ is defined as sign$(x)$ for $x\neq0$, and 0 for $x=0$. This workaround may be effective in controlling the energy of the parameters, but subgradient iterates do not attain zero, thus sparsification fails. Moreover, subgradient methods are substantially slow and may result in oscillations. These drawbacks are illustrated in the numerical example in Section \ref{sub:ne}.

Recently, proximal operator methods are used instead of subgradient, see, e.g., \cite{yan20,yun21}. However, their application and convergence proof are critical and limited to some specific cases.

Given these considerations, the proposed relaxed binary mask turns out to be an effective alternative strategy to match the use of non-differentiable, sparsity-promoting regularizers with standard gradient-based training algorithms.

\section{Theoretical analysis}\label{sec:ana}
In this section, we prove theoretical results that support the effectiveness of the proposed method, by providing conditions that guarantee to extract an optimal sparse subnetwork topology. In particular, these results explain why a $\log$ regularizer may be preferable to $\ell_1$. Finally, we show an illustrative example that corroborates the theoretical findings.

As a thorough analysis is quite complex, we focus on the following problem: we assume that a vector $\w$ of trained parameters is available and that a sparse mask $\mtrue\in\{0,1\}^d$ exists, such that $\mtrue \odot\w$ is a sparse topology with no substantial performance loss. Our aim is to estimate this optimal $\mtrue$. To this purpose, we solve
\begin{equation}\label{eq:subproblem}
\begin{split}
\min_{\m\in[0,1]^d}& \loss(x; \m\odot \w)+\lambda\Reg(\m)\\
\end{split}
\end{equation}
where $\Reg(\m)$ satisfies Assumption \ref{ass:reg}.

On the one hand, this problem is addressed in the context of the strong lottery ticket hypothesis, where subnetworks are extracted from randomly weighted neural networks without modifying the weight values, see \cite{ram20,mal20}. On the other hand, if in Eq.~\eqref{eq:theproblem} we proceed by alternated minimization over $\w$ and $\m$, Eq.~\eqref{eq:subproblem} can be interpreted as a sub-problem of \eqref{eq:theproblem}.


Since, in this section, we consider $\m$ as the unique variable, for simplicity we write $\loss(x; \m\odot \w)=\loss(\m)$.
Let 
\begin{equation}\label{mstar}
    \mstar=\argmin{\m\in[0,1]^d} \loss(\m)+\lambda\Reg(\m).
\end{equation}
Then, $\mstar$ is an estimate of $\mtrue$. To evaluate the accuracy of this estimate, we analyze the distance between $\mstar$ and $\mtrue$.
To this end, we assume that
\begin{equation}\label{max_gradiente}
    \|\nabla\loss(\mtrue)\|_{\infty}\leq \lambda.
\end{equation}
This is always true by properly choosing $\lambda$. On the other hand, we expect  $\|\nabla\loss(\mtrue)\|_{\infty}$ to be small as far as $\mtrue$ is close to a stationary point, i.e., if the model is amenable to effective sparsification.

To keep the analysis straightforward, we do the following convexity assumption.
\begin{assumption}\label{ass:convex}
There exists $\gamma>0$ such that \begin{equation}\label{lowerbound}
    \loss(\mstar)\geq \loss(\mtrue)+\nabla \loss(\mtrue)^T(\mstar-\mtrue)+\frac{\gamma}{2}\|\mstar-\mtrue\|_2^2.
\end{equation}
\end{assumption}
In other terms, we require strong convexity at least with respect to the points $\mtrue$ and $\mstar$. Clearly, this is fulfilled if the loss is globally strongly convex.
\subsection{Accuracy analysis for $\ell_1$ regularization}
In this section, we study the accuracy, in terms of distance between $\mtrue$ and $\mstar$, in the case of $\ell_1$ regularization.
Specifically, we prove the following result.
\begin{theorem}\label{theo:1}
Let $h=\mstar-\mtrue$. If $\Reg(\m)=\|\m\|_1$, then
\begin{equation}\label{bound:theo1}
     \|h\|_2\leq \frac{4\lambda\sqrt{k}}{\gamma}.
\end{equation}
\end{theorem}
\proof 
By definition of $\mstar$ in \eqref{mstar}, we have
\begin{equation}
   \loss(\mstar)+\lambda\Reg(\mstar)\leq \loss(\mtrue)+\lambda\Reg(\mtrue)
\end{equation}
Let $\supp$ and $\suppc$ denote the support of $\mtrue$ and its complementary set, respectively. Moreover, $h_{\supp}$ (respectively, $h_{\suppc}$) is the subvector of $h$ with components indexed in $\supp$ (respectively, in $\suppc$).  Then,
\begin{equation}\label{upperbound}
\begin{split}
    \loss(\mstar)-\loss(\mtrue)&\leq \lambda\|\mtrue\|_1-\lambda\|\mstar\|_1\\
    &=\lambda\left[\|\mtrue\|_1-\|\mtrue+h\|_1\right]\\
    &=\lambda\left[\|\mtrue\|_1-\|\mtrue_{\supp}+h_{\supp}\|_1-\|h_{\suppc}\|_1\right]\\
    &\leq\lambda\left[\|\mtrue\|_1-\|\mtrue_{\supp}\|+\|h_{\supp}\|_1-\|h_{\suppc}\|_1\right]\\
    &= \lambda\|h_{\supp}\|_1-\lambda\|h_{\suppc}\|_1.
\end{split}
\end{equation}
Moreover,
\begin{equation}\label{boundgrad}
\begin{split}
    \nabla \loss(\mtrue)^T(\mstar-\mtrue)&\leq\|\nabla \loss(\mtrue)\|_{\infty}^T\|h\|_1\\
    &\leq\lambda\|h\|_1.
\end{split}
\end{equation}
Now, from \eqref{lowerbound} and \eqref{upperbound}, we have a lower bound and an upper bound for $\loss(\mstar)-\loss(\mtrue)$. By using also \eqref{boundgrad}, we obtain
\begin{equation}
-\lambda \|h\|_1+\frac{\gamma}{2}\|h\|_2^2\leq \loss(\mstar)-\loss(\mtrue) \leq \lambda\|h_S\|_1-\lambda\|h_{\suppc}\|_1.
\end{equation}
Then,
\begin{equation}
\begin{split}
\frac{\gamma}{2}\|h\|_2^2 &\leq \lambda \|h\|_1+\lambda\|h_S\|_1-\lambda\|h_{\suppc}\|_1\\  &\leq \lambda \|h_S\|_1+\lambda \|h_{\suppc}\|_1+\lambda\|h_S\|_1-\lambda\|h_{\suppc}\|_1\\
&\leq 2\lambda \|h_S\|_1\\
&\leq 2\lambda \sqrt{k}\|h\|_2.
\end{split}
\end{equation}
Then, by dividing by $\|h\|_2$, we obtain the claim.\qed

The statement of Theorem \ref{theo:1} provides an error bound on the estimate $\mstar$ of the optimal mask $\mtrue$. This error bound is proportional to $\lambda$, which can be chosen as equal to $\|\nabla\loss(\mtrue)\|_{\infty}$ based on \eqref{max_gradiente}. In other terms, the more the masked solution $\mtrue\odot\theta$ is close to the non-masked optimum $\theta$, the more by solving the $\ell_1$-regularized problem we obtain a good estimate. Similarly, the more sparsity $k$ of $\mtrue$ is small, the more $\mstar$ is a good estimate. The third parameter present in the error bound is $\gamma$, which is a measure of the convexity of the loss in Assumption \ref{ass:convex}. Then, the error bound is smaller if the loss is more convex, i.e., $\gamma$ is larger.
An enhancement of Theorem \ref{theo:1} can be obtained by replacing $\ell_1$ with a more general strictly concave regularization. 
\subsection{Accuracy analysis for strictly concave  regularization}
In this section, we analyse the accuracy in case of strictly concave regularization.
\begin{theorem}\label{theo:2}
Let $h=\mstar-\mtrue$. Let $\Reg(\m)$ be strictly concave in $[0,1]$ as in Assumption \ref{ass:reg}.
Then
\begin{equation}
      \|h\|_2\leq \frac{4\lambda\sqrt{k}}{\gamma}-\phi(\mstar)
\end{equation}
where $\phi(\mstar)>0$ is assessed in the proof.
\end{theorem}
\proof 
\begin{equation}\label{upperbound_concave}
\begin{split}
    \loss(\mstar)-\loss(\mtrue)&\leq \lambda\Reg(\mtrue)-\lambda\Reg(\mstar).
\end{split}
\end{equation}
Now, since $\mtrue\in\{0,1\}^d$, then $\Reg(\mtrue)=\|\mtrue\|_1=\sum_{\mtrue_i}$.
Then,
\begin{equation}
\begin{split}
    \Reg(\mtrue)-\Reg(\mstar)&=\|\mtrue\|_1-\Reg(\mstar)\\
    &=\|\mtrue\|_1-\Reg(\mstar)\pm \|\mstar\|_1\\
    &\leq \|h_S\|_1-\|h_{\suppc}\|_1 -\phi(\mstar)
\end{split}
\end{equation}
where
\begin{equation}
    \phi(\mstar) =\Reg(\mstar)-\|\mstar\|_1\geq 0.
\end{equation}
\qed

Theorem \ref{theo:2} states that by replacing $\ell_1$ with a strictly concave regularization, such as the logarithmic one,  we obtain an error bound which is smaller than the one in Theorem \ref{theo:1} of $\phi(\mstar)$.

Let us discuss more in detail the role of $\phi(\mstar)$. We notice that if $\mstar\in\{0,1\}^d$, then $\Reg(\mstar)=\|\mstar\|_1$, $\phi(\mstar)=0$, and no enhancement is obtained by using a strictly concave regularizer. However, we know from the statements of Theorems \ref{theo:1} and \ref{theo:2} that $\|\mstar-\mtrue\|_2\leq \frac{4\sqrt{k}}{\gamma}\lambda$. Then,
\begin{lemma}
If $\frac{4\sqrt{k}}{\gamma}\lambda<1$ and $\mstar\in\{0,1\}^d$, then $\mstar=\mtrue$.
\end{lemma}
Therefore, if $\frac{4\sqrt{k}}{\gamma}\lambda<1$, the unique feasible binary solution is $\mtrue$. Moreover, if $\mstar\notin\{0,1\}^d$, then $\phi(\mstar)>0$ and we obtain a smaller error bound by using a strictly concave regularizer.
\subsection{Illustrative example}\label{sub:ne}
To conclude the analysis, we show an illustrative example of Problem \eqref{eq:subproblem} with Assumption \ref{ass:convex}. Specifically, we consider a problem of binary classification performed through logistic regression. We use the MNIST dataset \cite{MNIST} restricted to the digits 0 and 1. Each image is composed of $d=400$ pixels. 
 We consider 200 samples for each digit, 160  for training, and $N=40$ for validation test.
The corresponding cross-entropy loss function is
\begin{equation*}
    \loss(X,y; \w)=\sum_{i=1}^N y_i\log(\yhat_i)-(1-y_i)\log(1-\yhat_i)
\end{equation*}
where $y\in\{0,1\}^N$ and $\yhat\in[0,1]^N$ respectively are the vectors of correct and estimated labels. More precisely,  $$\yhat_i=\frac{1}{1+e^{X_i\w}}$$ where $\w\in\R^d$ is the vector of estimated parameters and $X\in\R^{N,401}$ is the validation dataset (including the intercept). The loss is convex; if an $\ell_2$ regularization is added, we have strong convexity as in Assumption \ref{ass:convex}.

\begin{figure}[t]
   \centering
    \includegraphics[width=0.7\columnwidth]{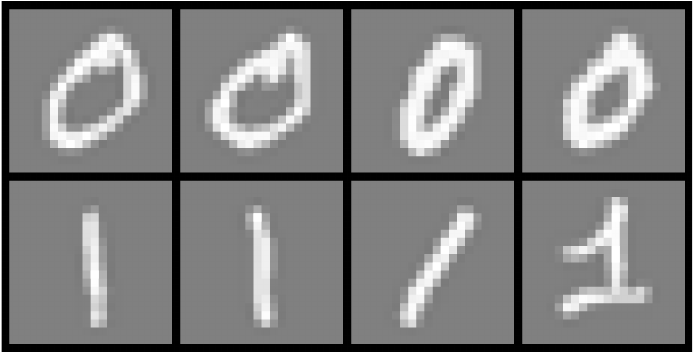}
     \caption{Eight samples from the MNIST dataset}
     \label{fig:mnist}
\end{figure}

As we can see in Fig. \ref{fig:mnist}, many pixels (e.g., the ones towards the image borders) are not significant for classification, therefore there is room for sparsification. Then, we tested different approaches for sparsification. First, we considered a classic $\ell_1$ regularization, i.e., we minimize  $\loss(X,y; \w)+\lambda \|\w\|_1$.  Secondly, we applied $\ell_1$ and logarithmic regularizations to the mask. More precisely, given a vector $\what$ of parameters, e.g., obtained via logistic regression, we aim at finding a binary mask $m\in\{0,1\}^d$ that selects the significant pixels, i.e., the final parameter vector will be $\what\odot m$.
Since $X\what\odot m =X\diag(\what)m$, for finding $m$ it is sufficient to train over the dataset $X\diag(\what)$, with the chosen regularization. In formulas, we minimize 
\begin{equation*}
    \loss(X\diag(\what),y; \m)+\lambda\Reg(\m),~~\m\in[0,1]^d
\end{equation*}
where $\Reg$ is either $\ell_1$ or log regularization. We notice that this approach does not retrain the model, but it just performs a selection of the parameters/pixels that can be neglected.

In the experiment, we do not set the ground truth $\mstar$, but we consider acceptable solutions whose accuracy is comparable to the one of standard logistic regression.

\begin{figure}[t]
    \centering
    \includegraphics[width=0.48\textwidth]{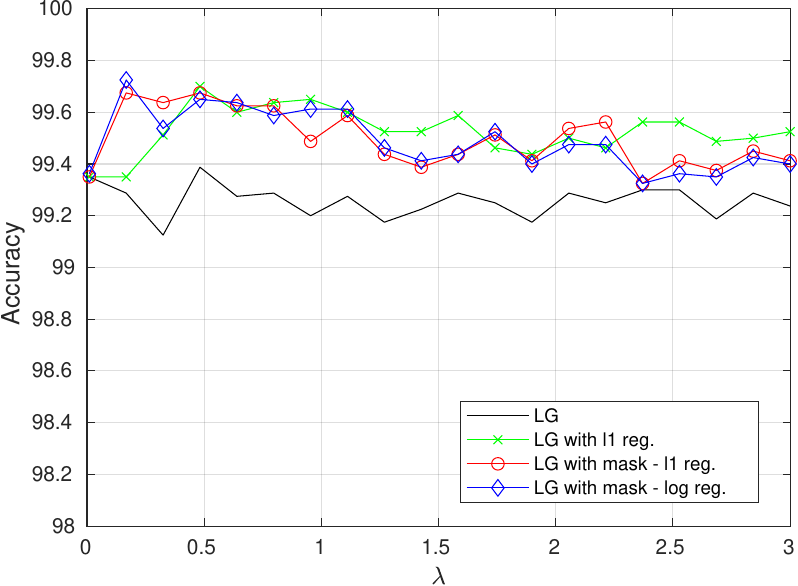}
        \includegraphics[width=0.48\textwidth]{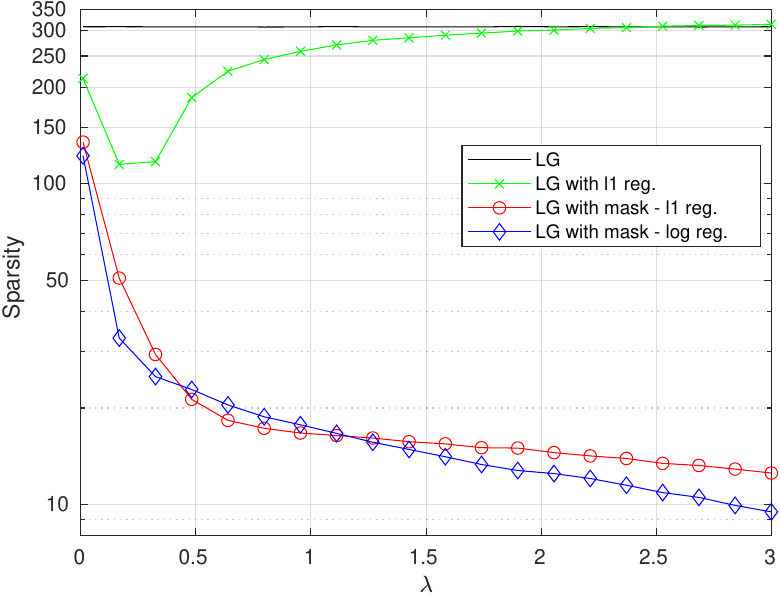}
    \caption{Binary classification on MNIST dataset with logistic regression (LG): accuracy and sparsity with respect to the design parameter $\lambda$. 
    }
    \label{fig:lg}
\end{figure}
\begin{figure}[h]
   \centering
    \begin{subfigure}[b]{1\columnwidth}
         \centering
    \includegraphics[width=0.3\columnwidth]{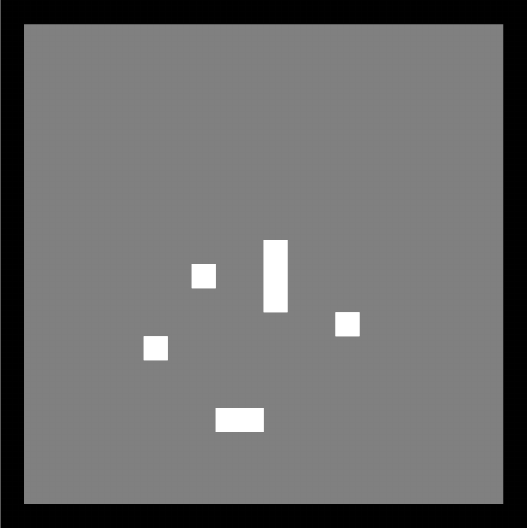}
    \caption{}
   \end{subfigure}
   \vskip0.1cm
   \begin{subfigure}[b]{1\columnwidth}
         \centering
   \includegraphics[width=0.7\columnwidth]{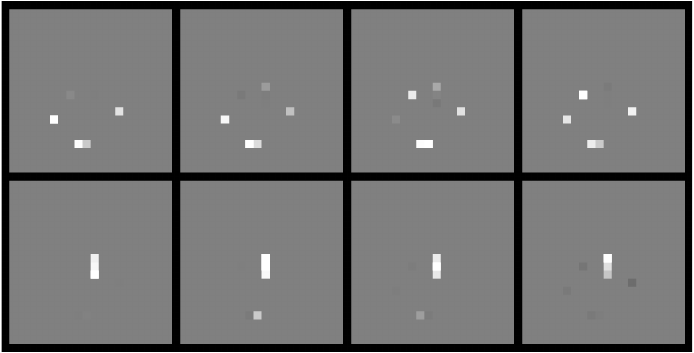}
   \caption{}
    \end{subfigure}
    \caption{(a) An 8-sparse mask obtained with logarithmic regularization; (c) Samples of Fig. \ref{fig:mnist}  masked via (b). We can see that 8 pixels may be sufficient to distinguish between digits 0 and 1.}
    \label{fig:mask_mnist}
\end{figure}

The results of the experiment can be visualized in Fig. \ref{fig:lg}, where we depict the mean accuracy and sparsity over 100 random runs, with respect to different values of the design parameter $\lambda$. In all the approaches, the minimization is performed via gradient descent. More precisely, for the $\ell_1$ approach on $\w$, a subgradient method is used given the non-differentiability of $\ell_1$. As previously discussed, this may not provide sparse solutions, which is clearly visible in the experiment: basically, a larger $\lambda$ causes large oscillations around zero and the solution is not sparse. This makes this solution not useful, and even less sparse than the original logistic regression. This issue is solved by using the mask: since the relaxed mask is positive we can use a projected gradient descent, which yields sparse solutions.

In particular, we notice that all the methods achieve comparable accuracy, but the proposed mask $\ell_1$/log methods efficiently sparsify the model. As expected from the theoretical results, in most cases the logarithmic version is more accurate than the $\ell_1$ version, by providing sparser models.

In this regard, in Fig. \ref{fig:mask_mnist}-(a), we can see an 8-sparse mask obtained via logarithmic regularization. By applying this mask on the eight digits in Fig. \ref{fig:mnist}, we obtain the masked digits in \ref{fig:mask_mnist}-(b): even though extremely sparsified, from visual inspection, we can verify that 0 and 1 are distinguishable.  

 \section{Experiments}\label{sec:exp}
 In this section, we perform an experimental evaluation of the proposed method. We first illustrate the experimental settings, then we discuss the experimental performance of the proposed method both in the context of the lottery ticket hypothesis and in network pruning. Then, an ablation study is conducted to evaluate the impact of the various aspects of the proposed technique. Finally, we report the training times of our method.
 \subsection{Concave regularizers}
 In the experimental evaluation we consider two concave regularizers, chosen among the most popular ones. The first one is $~\reg_{1}(m_i)=\m_i$. As discussed in the previous section, this regularizer corresponds to the $\ell_1$-norm restricted to the interval $[0,\ 1]$ and represents the limit case since it is both convex and concave. The second concave regularizer considered in the experiments is $~\reg_{\epsilon}(m_i)=\frac{\log\left(\frac{\m_i+\epsilon}{\epsilon}\right)}{\log\left(\frac{1+\epsilon}{\epsilon}\right)}$, which is strictly convex. 
 
 \subsection{Networks and datasets} 
 We test the chosen concave regularizers on various datasets and architectures. In particular, we focus on image classification. In this context, we consider three architectures: ResNet-20, VGG-11, and WideResNet-20 with 64 convolutional filters in the first block of the network. All these architectures are tested on CIFAR-10, CIFAR-100 and Tiny ImageNet datasets, resulting in nine distinct combinations. 
 In the context of network pruning, we evaluate the performance of the proposed method on CIFAR-10 using the ResNet-56 architecture, which is a standard experiment in this framework.  
 \subsection{Sparsities} 
 As discussed by \cite{fra20}, sparsities can be divided into three ranges. Trivial sparsities are the lowest sparsities, where even random pruning can achieve full accuracy. Matching sparsities correspond to moderate sparsities, where benchmark methods can approximately reach the accuracy of the full network. Extreme sparsities are the highest sparsities, where the subnetworks cannot reach full accuracy. In these experiments, we consider matching sparsities and the lowest extreme sparsities.

 
 \subsection{Experimental settings} 
 For all the methods considered in the experimental evaluation, we perform 3 pruning rounds to identify the sparse subnetwork. In each round, we follow the standard setup proposed in \cite{fra19}, training with SGD, a learning rate of 0.1, a weight decay of 0.0001, and a momentum of 0.9. Each round constitutes 85 epochs for ResNet-20 and 120 for VGG-11 and WideResNet-20, using a batch size of 128. We decay the learning rate by a factor of 10 at epochs 56 and 71 for ResNet-20 and epochs 60 and 90 for VGG-11 and WideResNet-20. The results are averaged on three runs, with different random seeds. For the concave regularizers, we set $\lambda=3\times 10^{-6}$ for $~\reg_{1}(m_i)$ and $\lambda=10^{-6}$ for $~\reg_{\epsilon}(m_i)$. In addition, for $\reg_{\epsilon}(m_i)$ we set $\epsilon=0.1$. All the methods considered in the experiments were implemented using the publicly available open\_lth framework\footnote{https://github.com/facebookresearch/open\_lth} (MIT license). All the experiments were conducted using two NVIDIA TITAN Xp GPUs. 
  \begin{figure}[t]
    \centering
    \includegraphics[width=0.45\textwidth]{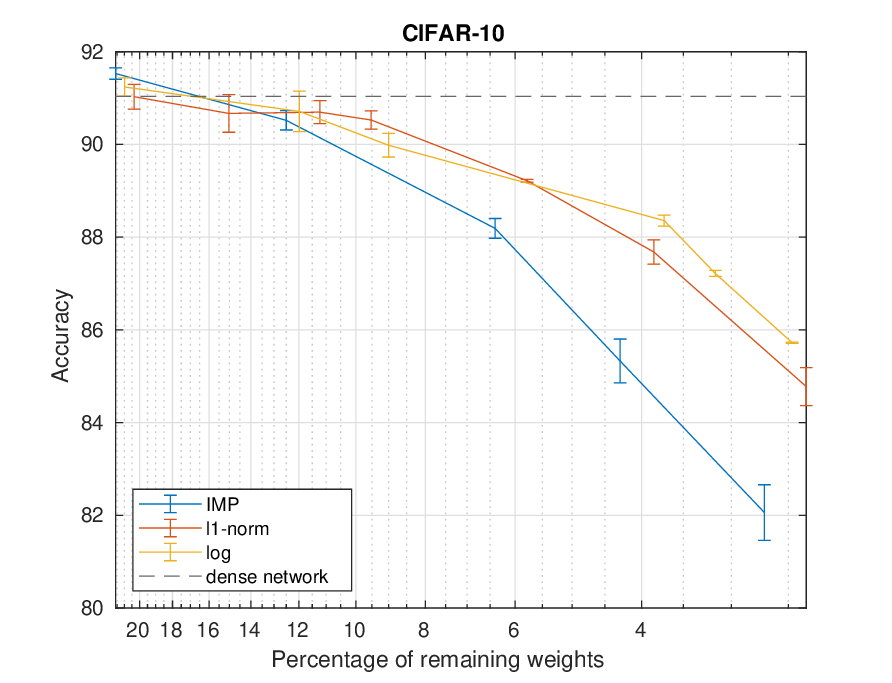}
    \includegraphics[width=0.45\textwidth]{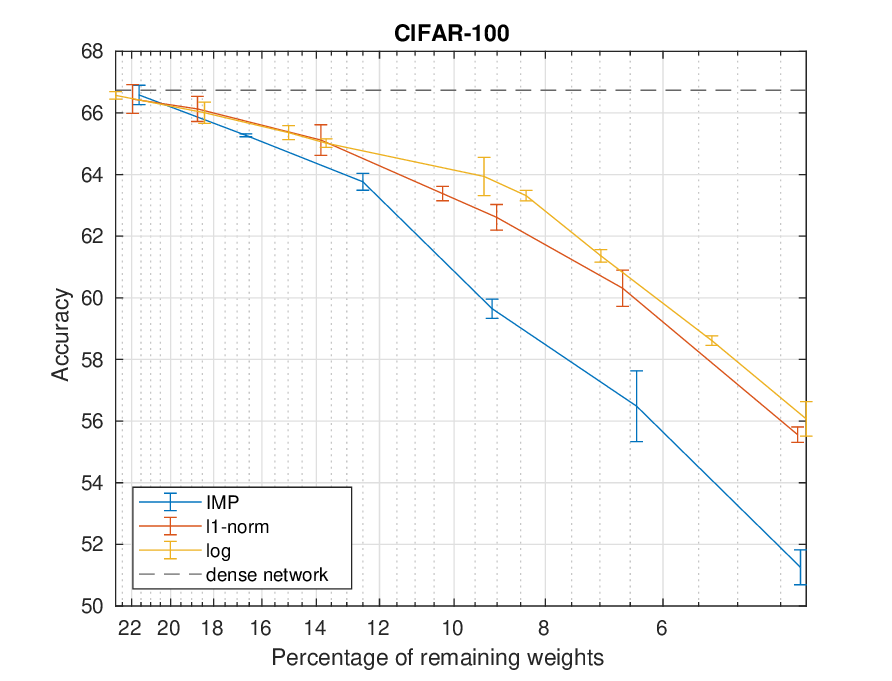}
    \includegraphics[width=0.45\textwidth]{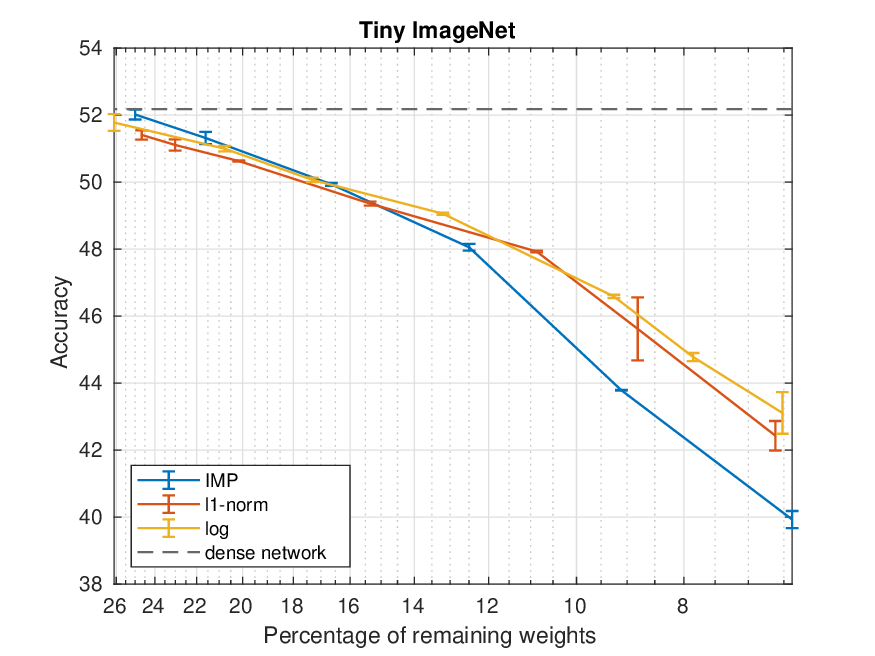}
    \caption{Test accuracy and sparsity of the subnetworks of ResNet-20 produced by IMP and the proposed method on CIFAR-10, CIFAR-100, and Tiny ImageNet.}
    \label{fig:res_resnet}
\end{figure}
  
 \begin{figure}[t]
    \centering
    \includegraphics[width=0.45\textwidth]{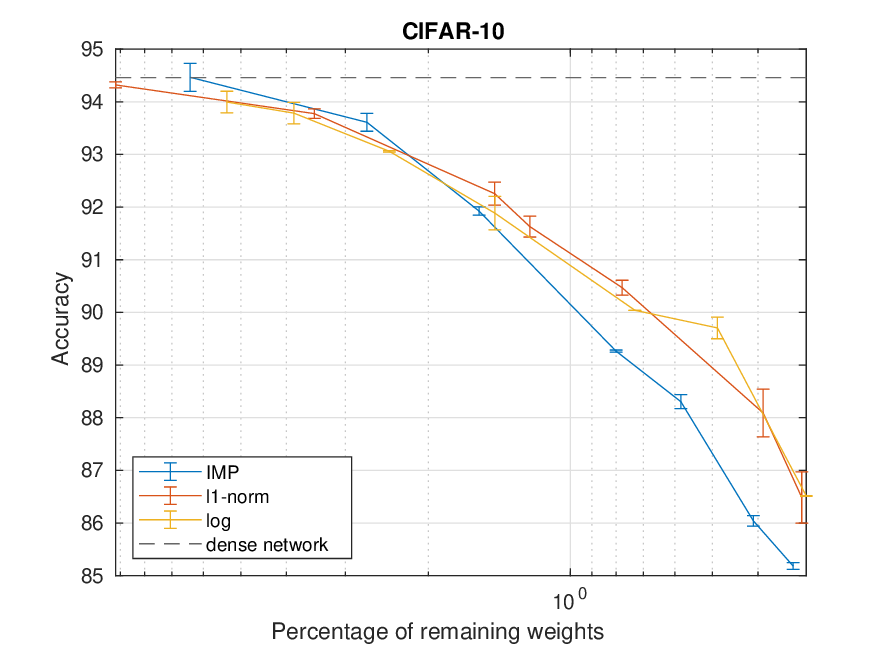}
    \includegraphics[width=0.45\textwidth]{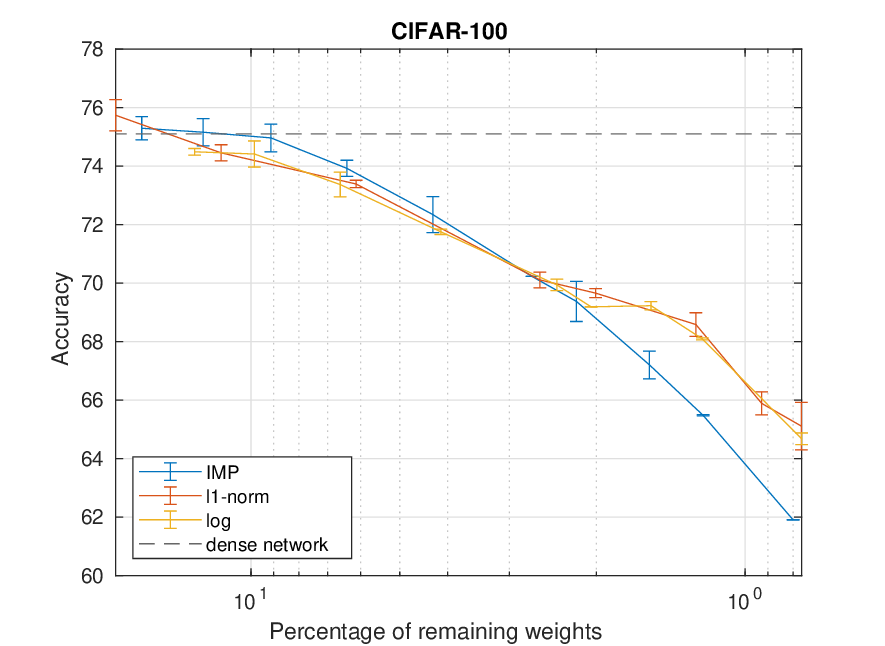}
    \includegraphics[width=0.45\textwidth]{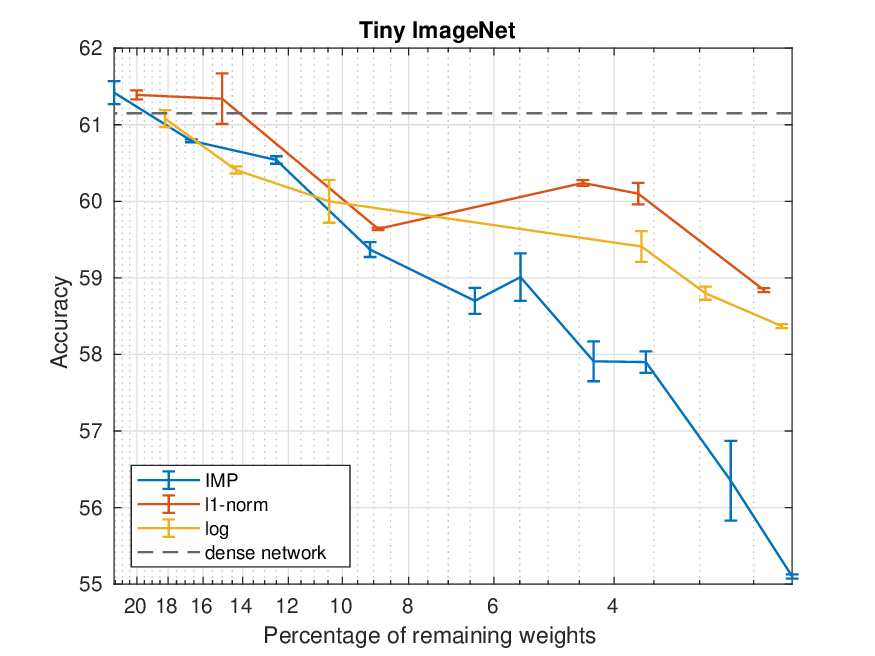}
    \caption{Test accuracy and sparsity of the subnetworks of WideResNet-20 produced by IMP and the proposed method on CIFAR-10, CIFAR-100, and Tiny ImageNet.}
    \label{fig:res_wideresnet}
\end{figure}
\begin{figure}[t]
    \centering
    \includegraphics[width=0.45\textwidth]{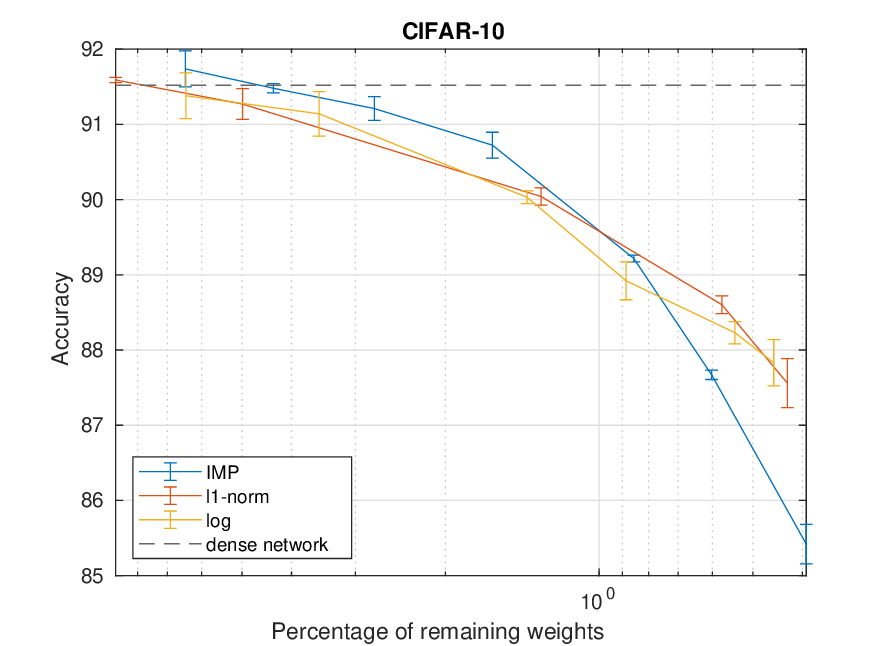}
    \includegraphics[width=0.45\textwidth]{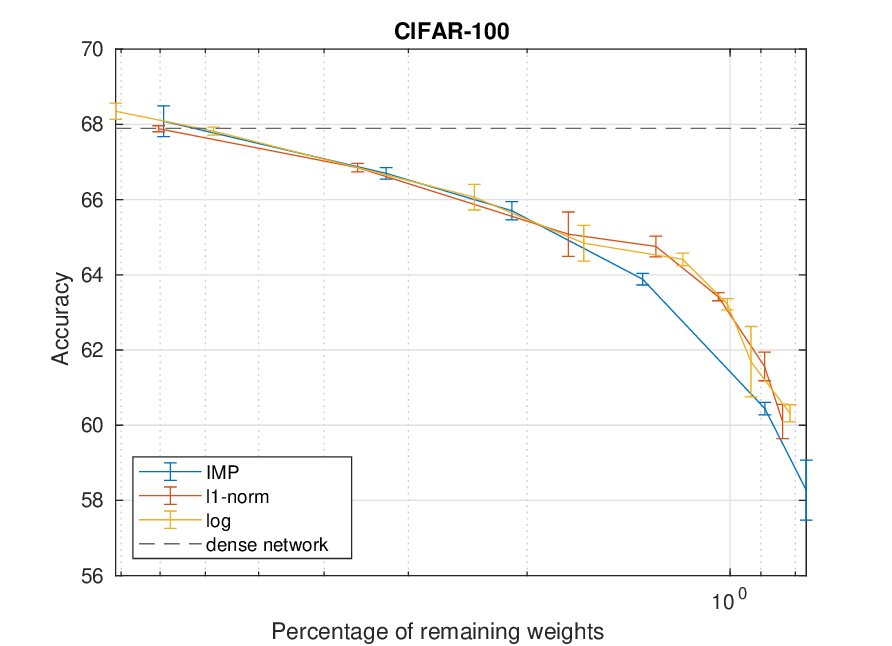}
    \includegraphics[width=0.45\textwidth]{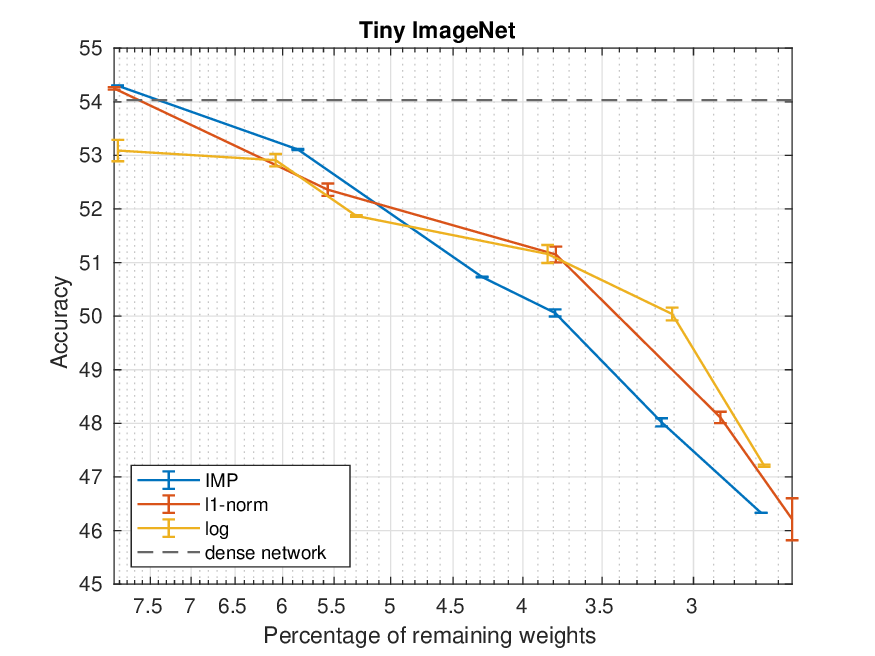}
    \caption{Test accuracy and sparsity of the subnetworks of VGG-11 produced by IMP and the proposed method on CIFAR-10, CIFAR-100, and Tiny ImageNet.}
    \label{fig:res_vgg11}
\end{figure}

 \subsection{Performance comparison}
We consider IMP as a benchmark method. Both the proposed method and IMP are implemented using the same training parameters described in the previous section. Fig. \ref{fig:res_resnet}, \ref{fig:res_wideresnet} and \ref{fig:res_vgg11} shows the results for ResNet-20, WideResNet-20, and VGG-11, respectively. We can observe that the proposed method with the concave regularizers reaches performance comparable to the ones of IMP at matching sparsities, but significantly outperforms IMP at higher sparsity, especially on the ResNet-20 architecture. The two concave regularizers exhibit similar performance in most of the experiments. However, in some cases, such as the experiments on ResNet-20 depicted in Fig. \ref{fig:res_resnet}, the strictly convex regularizer $~\reg_{\epsilon}(m_i)$ can significantly improve the performance with respect to $~\reg_{1}(m_i)$.

\subsection{Comparison against state-of-the-art pruning methods}
Even though the main focus of this work is the lottery ticket hypothesis where we aim to find sparse trainable networks, we also compare our work against state-of-the-art pruning methods that consider a dense-to-sparse approach. In this case, we evaluate the performance of the proposed method using a standard experiment for pruning methods, i.e., image classification on the CIFAR10 dataset using the ResNet-56 architecture. Results in Table~\ref{table:soa} show that our method greatly outperforms competing methods: we are able to improve over the baseline accuracy with an unpruned ratio as low as \(23\%\), which indicates that more than \(77\%\) of the parameters have been pruned.
%
\begin{table}[t]
 \centering
 \caption[]{Comparison against state-of-the-art pruning methods on the CIFAR10 dataset using the ResNet-56 architecture. Columns respectively report the baseline evaluation accuracy, the rate of unpruned parameters, the evaluation accuracy of the pruned model, and the accuracy difference against the baseline.}

\resizebox{\columnwidth}{!}{
\begin{tabular}{ c | c | c | c | c }
Method & Baseline (\%) & Unpruned (\%) & Acc. (\%) & Acc. Diff. (\%)  \\ 
\hline
GBN-40 \cite{you2019gate} & 93.10 & 46.50 &  93.43 & +0.33 \\
GBN-30 \cite{you2019gate} & 93.10 & 33.30 &  93.07 & -0.03 \\
Li et. al \cite{li2016pruning} & 93.10 & 86.30 & 93.08 & -0.02 \\
NISP \cite{yu2018nisp} & 93.10 & 57.40 &  93.13 & +0.03 \\
DCP-A \cite{zhuang2018discrimination} & 93.10 & 29.70 &  93.09 & -0.01 \\
SCOP \cite{tang2020scop} & 93.70 & 43.70 &  93.64 & -0.06 \\
SFP \cite{he2018soft} & 93.59 & 49.40 &  92.26 & -1.33 \\
GAL \cite{lin2019towards} & 93.26 & 55.20 &  92.74 & -0.52 \\
GAL-0.6 \cite{lin2019towards} & 93.26 & 88.20 &  92.98 & -0.28 \\
FPGM \cite{he2019filter} & 93.59 & 49.4 &  93.49 & -0.10 \\
HRank \cite{lin2020hrank} & 93.59 & 57.60 &  93.50 & -0.09 \\
HRank \cite{lin2020hrank} & 93.59 & 31.90 &  90.72 & -2.87 \\
White-Box \cite{zha23} & 93.26 & 43.70 &  93.20 & -0.06 \\
AMC \cite{he2018amc} & 92.80 & 50.00 &  91.90 & -0.90 \\
SCP \cite{kang2020operation} & 93.69 & 48.50 &  93.23 & -0.46 \\
LFPC \cite{he2020learning} & 93.26 & 47.10 &  93.24 & -0.02 \\
DSA \cite{ning2020dsa} & 93.12 & 46.40 &  92.92 & -0.66 \\
E-Pruner \cite{lin2021network} & 93.18 & 71.36 &  93.10 & -0.08 \\
FilterSketch \cite{lin2021filter} & 93.26 & 79.40 &  93.65 & +0.39 \\
FilterSketch \cite{lin2021filter} & 93.26 & 58.80 &  93.19 & -0.07 \\
FilterSketch \cite{lin2021filter} & 93.26 & 28.20 &  91.20 & -2.06 \\
\textbf{Ours (l1-norm)}& 93.26 & 1.79 &  88.06 & -5.02 \\
\textbf{Ours (l1-norm)} & 93.26 & 5.36 &  92.18 & -1.08 \\
\textbf{Ours (l1-norm)} & 93.26 & 9.93 &  92.99 & -0.27 \\
\textbf{Ours (l1-norm)} & 93.26 & 19.34 &  93.13 & -0.13 \\
\textbf{Ours (l1-norm)} & 93.26 & 22.10 &  93.28 & 0.02 \\
\textbf{Ours (Log)} & 93.26 & 1.66 &  87.98 & -5.28 \\
\textbf{Ours (Log)} & 93.26 & 5.12 &  92.17 & -1.09 \\
\textbf{Ours (Log)} & 93.26 & 9.91 &  93.01 & -0.25 \\
\textbf{Ours (Log)} & 93.26 & 18.47 &  93.11 & -0.15 \\
\textbf{Ours (Log)} & 93.26 & 23.2 &  93.30 & 0.04 \\

\hline
\end{tabular}}
\label{table:soa}
\end{table}
\subsection{Ablation study}
 \begin{figure}
    \centering
    \includegraphics[width=0.45\textwidth]{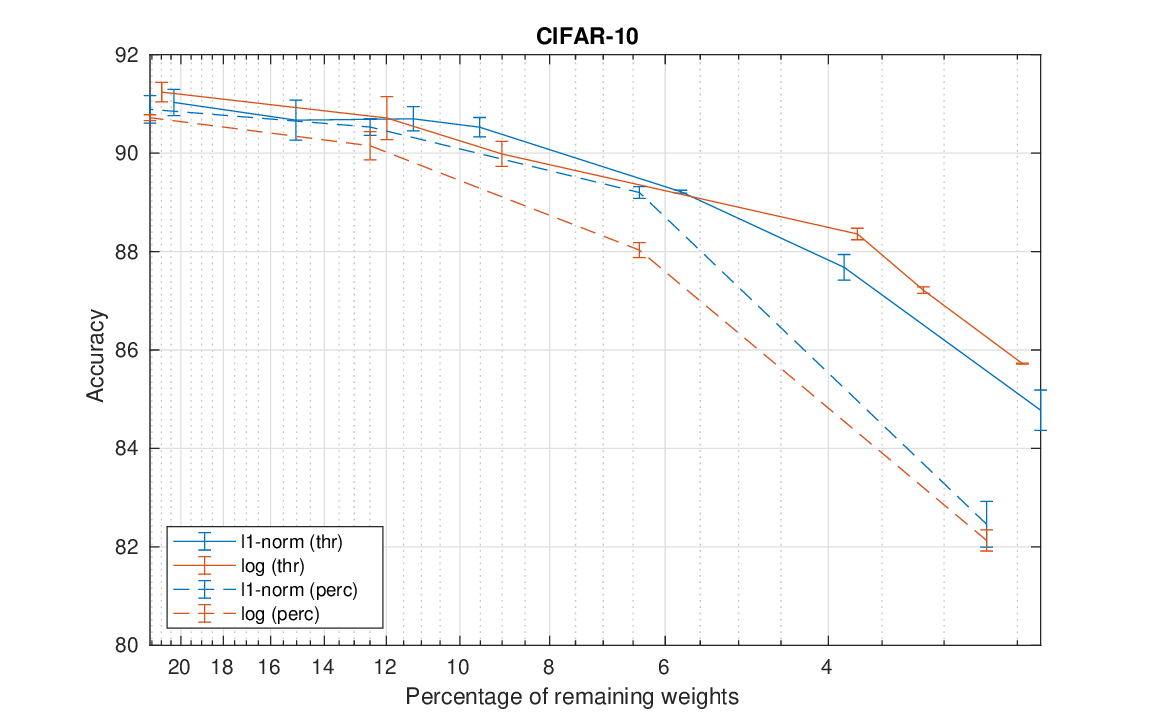}
    \caption{Test accuracy and sparsity of the subnetworks of ResNet-20 produced by the proposed method on CIFAR-10. Solid lines correspond to the case where we prune the weights whose mask values are below a given threshold (soft approach). Dashed lines correspond to the case where at each round we prune a fixed percentage of weights based on the magnitude of the mask value (hard approach).}
    \label{fig:ab1}
\end{figure}
\begin{figure}
    \centering
    \includegraphics[width=0.45\textwidth]{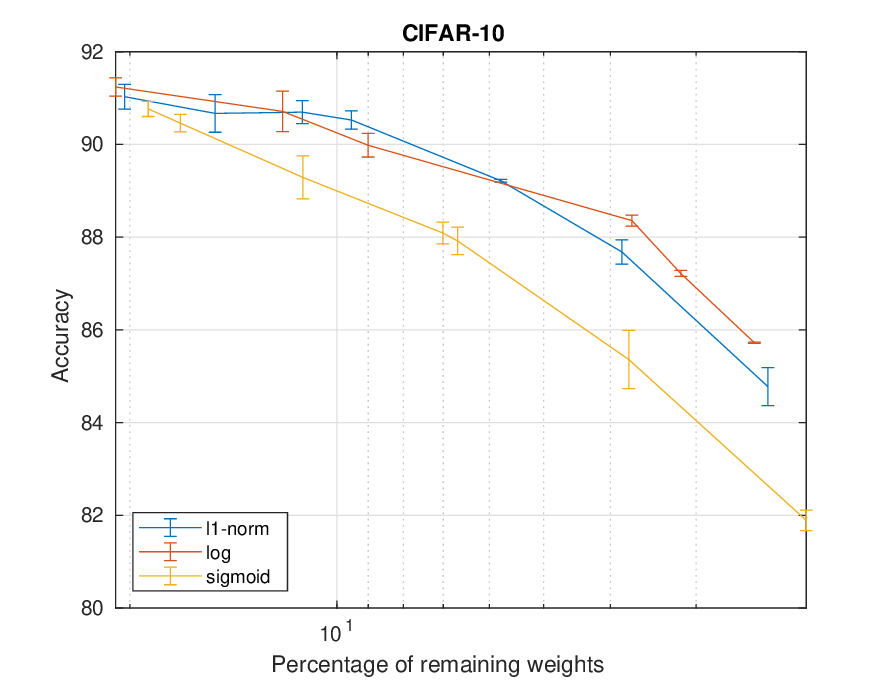}
    \includegraphics[width=0.45\textwidth]{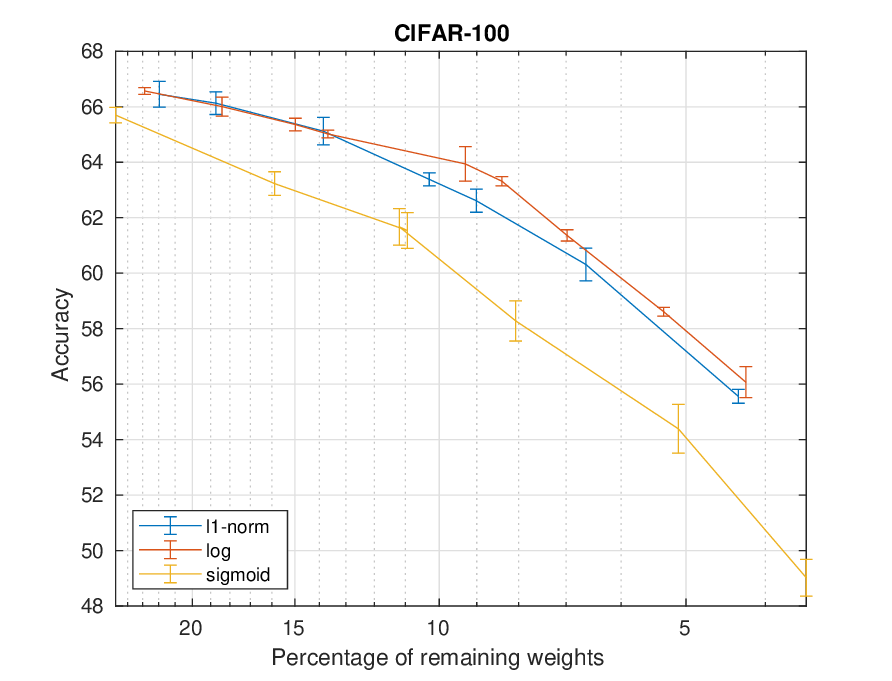}
    \caption{Test accuracy and sparsity of the subnetworks of ResNet-20 produced the proposed method using concave regularizer or sigmoid reparameterization on CIFAR-10 (left) and CIFAR-100 (right).}
    \label{fig:ab2}
\end{figure}
 In this section, we evaluate the impact of some aspects of the proposed method. First, we investigate the impact of setting a threshold instead of removing a fixed percentage of parameters. In the first case we remove the parameters whose mask values are below a given threshold, at the end of each round. In contrast, in the second case, a constant percentage of weights is removed at each round based on the magnitude of the mask values. Fig. \ref{fig:ab1} shows the results of this comparison for ResNet-20 on CIFAR-10. We can observe that setting a threshold can significantly improve the performance of the proposed method.
 
We then evaluate the importance of using concave regularizers, comparing them against a different regularization technique. We consider a sigmoid reparameterization  $\m=\sigma(\beta s)=\frac{1}{1+e^{-\beta s}}$, as proposed in \cite{sav20}. The parameter $\beta$ is defined as in \cite{sav20}, where is updated at each epoch using an exponential schedule $\beta^{(t)}=(\beta^{(T)})^{\frac{t}{T}}$ for epoch $t=0,...,T$, where we set $\beta=200$. The results for ResNet-20 on CIFAR-10 and CIFAR-100 are shown in Fig. \ref{fig:ab2}. We can observe that the proposed concave regularizers can provide significantly better performance than the sigmoid reparameterization. 
 \subsection{Training times}
 In Table \ref{table-traintime} we indicate the training times on CIFAR-10 of the architectures considered in the experiments. Table \ref{table-traintime} shows both the time for finding the sparse architecture and the training time of the sparse network.

 \begin{table}[t]
 \centering
\caption{Training times on CIFAR-10 of the architectures considered in the experiments.}\label{table-traintime}
{\small{
 \begin{tabular}{l|c|c}
 Architecture & \begin{tabular}{@{}c@{}}Time to find \\ the sparse network\end{tabular} & \begin{tabular}{@{}c@{}}Training time of \\ the sparse network\end{tabular} \\
 \hline
 ResNet-20 & 91 minutes & 27 minutes \\
 WideResNet-20 & 168 minutes & 54 minutes \\
 VGG-11 & 119 minutes & 40 minutes \\
 ResNet-56 & 244 minutes & 67 minutes \\

\end{tabular}}}
\end{table}
 \section{Conclusion}\label{sec:con}
In this paper, we proposed a novel iterative method for identifying matching tickets, i.e., for extracting subnetworks that, when trained in isolation, achieve accuracy comparable to the dense networks that contain them. The method is based on the introduction of a binary mask, which is relaxed to a continuous variable and penalized by concave regularization to promote its sparsity. The use of strictly concave regularization is quite novel in deep learning. We provide theoretical results that substantiate its effectiveness with respect to other methods. Experiments on different problems and architectures show that the proposed approach is valuable with respect to the iterative magnitude pruning method. Future work will be devoted to optimizing the tuning of the hyperparameters and to the reduction of the numerical complexity, by considering early-bird approaches and untrained models as in the strong lottery ticket framework.

\bibliographystyle{IEEEtran}  
\bibliography{refs}

\begin{thebibliography}{10}
\providecommand{\url}[1]{#1}
\csname url@samestyle\endcsname
\providecommand{\newblock}{\relax}
\providecommand{\bibinfo}[2]{#2}
\providecommand{\BIBentrySTDinterwordspacing}{\spaceskip=0pt\relax}
\providecommand{\BIBentryALTinterwordstretchfactor}{4}
\providecommand{\BIBentryALTinterwordspacing}{\spaceskip=\fontdimen2\font plus
\BIBentryALTinterwordstretchfactor\fontdimen3\font minus
  \fontdimen4\font\relax}
\providecommand{\BIBforeignlanguage}[2]{{%
\expandafter\ifx\csname l@#1\endcsname\relax
\typeout{** WARNING: IEEEtran.bst: No hyphenation pattern has been}%
\typeout{** loaded for the language `#1'. Using the pattern for}%
\typeout{** the default language instead.}%
\else
\language=\csname l@#1\endcsname
\fi
#2}}
\providecommand{\BIBdecl}{\relax}
\BIBdecl

\bibitem{lecun90}
Y.~LeCun, J.~Denker, and S.~Solla, ``Optimal brain damage,'' in \emph{Proc.
  Adv. Neural Inform. Process. Syst. (NeurIPS)}, vol.~2, 1990.

\bibitem{han15}
S.~Han, J.~Pool, J.~Tran, and W.~Dally, ``Learning both weights and connections
  for efficient neural network,'' in \emph{Proc. Adv. Neural Inform. Process.
  Syst. (NeurIPS)}, 2015, pp. 1135--1143.

\bibitem{lou18}
C.~Louizos, M.~Welling, and D.~P. Kingma, ``Learning sparse neural networks
  through $\ell_0$ regularization,'' in \emph{Proc. Int. Conf. Learn.
  Represent. (ICLR)}, 2018.

\bibitem{fra19}
J.~Frankle and M.~Carbin, ``The lottery ticket hypothesis: Finding sparse,
  trainable neural networks.'' in \emph{Proc. Int. Conf. Learn. Represent.
  (ICLR)}, 2019.

\bibitem{yang2017designing}
T.-J. Yang, Y.-H. Chen, and V.~Sze, ``Designing energy-efficient convolutional
  neural networks using energy-aware pruning,'' in \emph{Proc. IEEE/CVF Conf.
  Comput. Vis. Pattern Recognit. (CVPR)}, 2017, pp. 5687--5695.

\bibitem{mol17}
P.~Molchanov, S.~Tyree, T.~Karras, T.~Aila, and J.~Kautz, ``Pruning
  convolutional neural networks for resource efficient inference,'' in
  \emph{Proc. Int. Conf. Learn. Represent. (ICLR)}, 2017.

\bibitem{blalock2020state}
D.~Blalock, J.~J. Gonzalez~Ortiz, J.~Frankle, and J.~Guttag, ``What is the
  state of neural network pruning?'' in \emph{Proc. Mach. Learning Syst.
  (MLSys}, vol.~2, 2020, pp. 129--146.

\bibitem{zhu18}
M.~H. Zhu and S.~Gupta, ``To prune, or not to prune: Exploring the efficacy of
  pruning for model compression,'' in \emph{Proc. Int. Conf. Learn. Represent.
  (ICLR)}, 2018.

\bibitem{yan20}
Y.~Yang, Y.~Yuan, A.~Chatzimichailidis, R.~J. van Sloun, L.~Lei, and
  S.~Chatzinotas, ``Prox{SGD}: Training structured neural networks under
  regularization and constraints,'' in \emph{Proc. Int. Conf. Learn. Represent.
  (ICLR)}, 2020.

\bibitem{yun21}
J.~Yun, A.~C. Lozano, and E.~Yang, ``Adaptive proximal gradient methods for
  structured neural networks,'' in \emph{Proc. Adv. Neural Inform. Process.
  Syst. (NeurIPS)}, vol.~34, 2021, pp. 24\,365--24\,378.

\bibitem{fra20}
J.~Frankle, G.~K. Dziugaite, D.~Roy, and M.~Carbin, ``Linear mode connectivity
  and the lottery ticket hypothesis,'' in \emph{Proc. Int. Conf. Mach. Learning
  (ICML)}, 2020, pp. 3259--3269.

\bibitem{sav20}
P.~Savarese, H.~Silva, and M.~Maire, ``Winning the lottery with continuous
  sparsification,'' in \emph{Proc. Adv. Neural Inform. Process. Syst.
  (NeurIPS)}, vol.~33, 2020, pp. 11\,380--11\,390.

\bibitem{liu17}
Z.~Liu, J.~Li, Z.~Shen, G.~Huang, S.~Yan, and C.~Zhang, ``Learning efficient
  convolutional networks through network slimming,'' in \emph{2017 IEEE
  International Conference on Computer Vision (ICCV)}, 2017, pp. 2755--2763.

\bibitem{liu19}
Z.~Liu, M.~Sun, T.~Zhou, G.~Huang, and T.~Darrell, ``Rethinking the value of
  network pruning,'' in \emph{Proc. Int. Conf. Learn. Represent. (ICLR)}, 2019.

\bibitem{tar22}
E.~Tartaglione, A.~Bragagnolo, F.~Odierna, A.~Fiandrotti, and M.~Grangetto,
  ``{SeReNe}: Sensitivity-based regularization of neurons for structured
  sparsity in neural networks,'' \emph{IEEE Trans. Neural Netw. Learn. Syst.},
  vol.~33, no.~12, pp. 7237--7250, 2022.

\bibitem{sal22}
H.~Salehinejad and S.~Valaee, ``{ED}ropout: Energy-based dropout and pruning of
  deep neural networks,'' \emph{IEEE Trans. Neural Netw. Learn. Syst.},
  vol.~33, no.~10, pp. 5279--5292, 2022.

\bibitem{ning2020dsa}
X.~Ning, T.~Zhao, W.~Li, P.~Lei, Y.~Wang, and H.~Yang, ``{DSA}: More efficient
  budgeted pruning via differentiable sparsity allocation,'' in \emph{Proc.
  Eur. Conf. Comput. Vis. (ECCV)}.\hskip 1em plus 0.5em minus 0.4em\relax
  Springer, 2020, pp. 592--607.

\bibitem{lin2020hrank}
M.~Lin, R.~Ji, Y.~Wang, Y.~Zhang, B.~Zhang, Y.~Tian, and L.~Shao, ``{HR}ank:
  Filter pruning using high-rank feature map,'' in \emph{Proc. IEEE/CVF Conf.
  Comput. Vis. Pattern Recognit. (CVPR)}, 2020, pp. 1529--1538.

\bibitem{lin2021network}
M.~Lin, R.~Ji, S.~Li, Y.~Wang, Y.~Wu, F.~Huang, and Q.~Ye, ``Network pruning
  using adaptive exemplar filters,'' \emph{IEEE Trans. Neural Netw. Learn.
  Syst.}, vol.~33, no.~12, pp. 7357--7366, 2021.

\bibitem{lin2021filter}
M.~Lin, L.~Cao, S.~Li, Q.~Ye, Y.~Tian, J.~Liu, Q.~Tian, and R.~Ji, ``Filter
  sketch for network pruning,'' \emph{IEEE Trans. Neural Netw. Learn. Syst.},
  vol.~33, no.~12, pp. 7091--7100, 2021.

\bibitem{zha23}
Y.~Zhang, M.~Lin, C.-W. Lin, J.~Chen, Y.~Wu, Y.~Tian, and R.~Ji, ``Carrying out
  {CNN} channel pruning in a white box,'' \emph{IEEE Trans. Neural Netw. Learn.
  Syst.}, vol.~34, no.~10, pp. 7946--7955, 2023.

\bibitem{tib96}
R.~Tibshirani, ``Regression shrinkage and selection via the {L}asso,'' \emph{J.
  Royal. Statist. Soc. B}, vol.~58, pp. 267--288, 1996.

\bibitem{can08rew}
E.~J. Cand\`es, M.~B. Wakin, and S.~Boyd, ``Enhancing sparsity by reweighted
  $\ell_1$ minimization,'' \emph{J. Fourier Anal. Appl.}, vol.~14, no. 5-6, pp.
  877--905, 2008.

\bibitem{woo16}
J.~Woodworth and R.~Chartrand, ``Compressed sensing recovery via nonconvex
  shrinkage penalties,'' \emph{Inverse Problems}, vol.~32, no.~7, pp.
  75\,004--75\,028, 2016.

\bibitem{zha10MCP}
C.-H. Zhang, ``Nearly unbiased variable selection under minimax concave
  penalty,'' \emph{Ann. Statist.}, vol.~38, no.~2, pp. 894--942, 2010.

\bibitem{fou09}
S.~Foucart and M.-J. Laui, ``Sparsest solutions of underdetermined linear
  systems via $\ell_q$ minimization for $0<q\leq 1$,'' \emph{Appl. Comput.
  Harmon. Anal.}, vol.~26, pp. 395--407, 2009.

\bibitem{sel17}
I.~Selesnick, ``Sparse regularization via convex analysis,'' \emph{IEEE Trans.
  Signal Process.}, vol.~65, no.~17, pp. 4481--4494, 2017.

\bibitem{fox20}
V.~Cerone, S.~M. Fosson, D.~Regruto, and A.~Salam, ``Sparse learning with
  concave regularization: relaxation of the irrepresentable condition,'' in
  \emph{IEEE Conf. Decis. Control (CDC)}, 2020, pp. 396--401.

\bibitem{wen18}
F.~Wen, L.~Chu, P.~Liu, and R.~C. Qiu, ``A survey on nonconvex
  regularization-based sparse and low-rank recovery in signal processing,
  statistics, and machine learning,'' \emph{IEEE Access}, vol.~6, pp.
  69\,883--69\,906, 2018.

\bibitem{bay16}
I.~Bayram, ``On the convergence of the iterative shrinkage/thresholding
  algorithm with a weakly convex penalty,'' \emph{IEEE Trans. Signal Process.},
  vol.~64, no.~6, pp. 1597--1608, 2016.

\bibitem{gon13}
P.~Gong, C.~Zhang, Z.~Lu, J.~Huang, and J.~Ye, ``A general iterative shrinkage
  and thresholding algorithm for non-convex regularized optimization
  problems,'' in \emph{Proc. Int. Conf. Mach. Learning (ICML)}.\hskip 1em plus
  0.5em minus 0.4em\relax PMLR, 2013, pp. 37--45.

\bibitem{bre15}
K.~Bredies, D.~A. Lorenz, and S.~Reiterer, ``Minimization of non-smooth,
  non-convex functionals by iterative thresholding,'' \emph{J. Optim. Theory
  Appl.}, vol. 165, no.~1, pp. 78--112, 2015.

\bibitem{hon16}
M.~Hong, Z.~Q. Luo, and M.~Razaviyayn, ``Convergence analysis of alternating
  direction method of multipliers for a family of nonconvex problems,''
  \emph{SIAM J. Optim.}, vol.~26, no.~1, pp. 337--364, 2016.

\bibitem{ram20}
V.~Ramanujan, M.~Wortsman, A.~Kembhavi, A.~Farhadi, and M.~Rastegari,
  ``What’s hidden in a randomly weighted neural network?'' in \emph{Proc.
  IEEE/CVF Conf. Comput. Vis. Pattern Recognit. (CVPR)}, 2020, pp.
  11\,890--11\,899.

\bibitem{mal20}
E.~Malach, G.~Yehudai, S.~Shalev-Schwartz, and O.~Shamir, ``Proving the lottery
  ticket hypothesis: Pruning is all you need,'' in \emph{Proc. Int. Conf. Mach.
  Learning (ICML)}, vol. 119, 2020, pp. 6682--6691.

\bibitem{MNIST}
L.~Deng, ``{The MNIST Database of Handwritten Digit Images for Machine Learning
  Research},'' \emph{IEEE Signal Processing Magazine}, vol.~29, no.~6, pp.
  141--142, 2012.

\bibitem{you2019gate}
Z.~You, K.~Yan, J.~Ye, M.~Ma, and P.~Wang, ``Gate decorator: Global filter
  pruning method for accelerating deep convolutional neural networks,''
  \emph{Proc. Adv. Neural Inform. Process. Syst. (NeurIPS)}, vol.~32, 2019.

\bibitem{li2016pruning}
H.~Li, A.~Kadav, I.~Durdanovic, H.~Samet, and H.~P. Graf, ``Pruning filters for
  efficient convnets,'' \emph{arXiv preprint arXiv:1608.08710}, 2016.

\bibitem{yu2018nisp}
R.~Yu, A.~Li, C.-F. Chen, J.-H. Lai, V.~I. Morariu, X.~Han, M.~Gao, C.-Y. Lin,
  and L.~S. Davis, ``Nisp: Pruning networks using neuron importance score
  propagation,'' in \emph{Proceedings of the IEEE conference on computer vision
  and pattern recognition}, 2018, pp. 9194--9203.

\bibitem{zhuang2018discrimination}
Z.~Zhuang, M.~Tan, B.~Zhuang, J.~Liu, Y.~Guo, Q.~Wu, J.~Huang, and J.~Zhu,
  ``Discrimination-aware channel pruning for deep neural networks,''
  \emph{Proc. Adv. Neural Inform. Process. Syst. (NeurIPS)}, vol.~31, 2018.

\bibitem{tang2020scop}
Y.~Tang, Y.~Wang, Y.~Xu, D.~Tao, C.~Xu, C.~Xu, and C.~Xu, ``Scop: Scientific
  control for reliable neural network pruning,'' \emph{Proc. Adv. Neural
  Inform. Process. Syst. (NeurIPS)}, vol.~33, pp. 10\,936--10\,947, 2020.

\bibitem{he2018soft}
Y.~He, G.~Kang, X.~Dong, Y.~Fu, and Y.~Yang, ``Soft filter pruning for
  accelerating deep convolutional neural networks,'' \emph{arXiv preprint
  arXiv:1808.06866}, 2018.

\bibitem{lin2019towards}
S.~Lin, R.~Ji, C.~Yan, B.~Zhang, L.~Cao, Q.~Ye, F.~Huang, and D.~Doermann,
  ``Towards optimal structured cnn pruning via generative adversarial
  learning,'' in \emph{Proceedings of the IEEE/CVF conference on computer
  vision and pattern recognition}, 2019, pp. 2790--2799.

\bibitem{he2019filter}
Y.~He, P.~Liu, Z.~Wang, Z.~Hu, and Y.~Yang, ``Filter pruning via geometric
  median for deep convolutional neural networks acceleration,'' in
  \emph{Proceedings of the IEEE/CVF conference on computer vision and pattern
  recognition}, 2019, pp. 4340--4349.

\bibitem{he2018amc}
Y.~He, J.~Lin, Z.~Liu, H.~Wang, L.-J. Li, and S.~Han, ``Amc: Automl for model
  compression and acceleration on mobile devices,'' in \emph{Proc. Eur. Conf.
  Comput. Vis. (ECCV)}, 2018, pp. 784--800.

\bibitem{kang2020operation}
M.~Kang and B.~Han, ``Operation-aware soft channel pruning using differentiable
  masks,'' in \emph{International Conference on Machine Learning}.\hskip 1em
  plus 0.5em minus 0.4em\relax PMLR, 2020, pp. 5122--5131.

\bibitem{he2020learning}
Y.~He, Y.~Ding, P.~Liu, L.~Zhu, H.~Zhang, and Y.~Yang, ``Learning filter
  pruning criteria for deep convolutional neural networks acceleration,'' in
  \emph{Proc. IEEE/CVF Conf. Comput. Vis. Pattern Recognit. (CVPR)}, 2020, pp.
  2009--2018.

\end{thebibliography}

\end{document}